\documentclass[preprint,12pt]{elsarticle}

\usepackage{amssymb}
\usepackage{amsmath,amssymb,amsfonts}
\usepackage{algorithmic}
\usepackage{graphicx}
\usepackage{textcomp}
\usepackage{amssymb}
\usepackage{amsmath}
\usepackage{tabularx}  
\usepackage{color}
\usepackage{makecell}
\usepackage{booktabs} 
\usepackage{booktabs,multirow}
\usepackage{graphicx} 
\usepackage{subfig} 
\usepackage{threeparttable}
\usepackage{caption}
\usepackage{xcolor}

\usepackage{amsmath}

\journal{Nuclear Physics B}

\begin{document}

\begin{frontmatter}



\title{Multiscale Dual-path Feature Aggregation Network for Remaining Useful Life Prediction of Lithium-Ion Batteries}


\author[1]{Zihao Lv\corref{cor1}}
\author[1]{Siqi Ai}
\author[2]{Yanbin Zhang}

\affiliation[1]{organization={Southwest University},
	addressline={College of Computer and Information Science}, 
	city={Chongqing},
	postcode={400715}, 
	state={Chongqing},
	country={China}}
	
\affiliation[2]{organization={Southwest University},
	addressline={College of Han Hong}, 
	city={Chongqing},
	postcode={408000}, 
	state={Chongqing},
	country={China}}

\cortext[cor1]{Corresponding author}

\begin{abstract}
Targeted maintenance strategies, ensuring the dependability and safety of industrial machinery. However, current modeling techniques for assessing both local and global correlation of battery degradation sequences are inefficient and difficult to meet the needs in real-life applications. For this reason, we propose a novel deep learning architecture, multiscale dual-path feature aggregation network (MDFA-Net), for RUL prediction. MDFA-Net consists of dual-path networks, the first path network, multiscale feature network (MF-Net) that maintains the shallow information and avoids missing information, and the second path network is an encoder network (EC-Net) that captures the continuous trend of the sequences and retains deep details. Integrating both deep and shallow attributes effectively grasps both local and global patterns. Testing conducted with two publicly available Lithium-ion battery datasets reveals our approach surpasses existing top-tier methods in RUL forecasting, accurately mapping the capacity degradation trajectory.
\end{abstract}


\end{frontmatter}



\section{Introduction}\label{sec1}

Lithium-ion batteries (LIBs) are core to electric vehicles, stationary storage, and portable electronics \cite{lv2024remaining, li2025learning, wang2024gt,chen2024sdgnn,luo2024pseudo,li2023nonlinear,chen2024state,yuan2022kalman,yuan2023adaptive,yuan2020multilayered,Wang2024,luo2020position,shang2021alpha,xin2019non,yuan2020generalized}. Their capacity, however, degrades with cycling, making accurate remaining useful life (RUL) prediction essential for safety, reliability, and cost-effective maintenance \cite{lv2025state, han2025sgd,yuan2020temporal,chen2021hyper,li2021proportional,yuan2018effects,he2024structure,wu2024outlier,wu2024online,tang2025neural,chen2024latent,wu2023robust}.

Existing approaches fall into three families \cite{liu2020remaining,liu2020cnn,chen2024robust,zeng2024novel,lv2023study,li2025semi,mei2024dual,WANG2025121315,WANG2023111129}. \textit{Direct measurement} (e.g., OCV, Coulomb counting) infers state-of-health (SOH) from current/voltage/impedance, but often requires long rest periods, suffers from integration drift, and is costly to deploy broadly \cite{mo2024semi,luo2021fast,wu2023graph,chen2023tensor, yuan2024node,lin20243d,yang2024latent,luo2021alternating,luo2022neulft}. \textit{Model-based} methods capture physicochemical degradation (e.g., SEI growth, active-material loss, plating) and can yield mechanistic insight, yet they require strong assumptions and struggle to generalize across usage profiles \cite{bi2023proximal, yuan2025proportional,liao2025local,wu2024fine,chen2024efficient,yang2024data,yang2021machine}. \textit{Data-driven} methods bypass explicit physics by mapping routine measurements to SOH/RUL via machine learning and deep networks \cite{yuan2024fuzzy, yuan2024adaptive, wang2021adaptive,zhong2024alternating,zhang2020deep,catelani2021remaining,li2020state,wang2024distributed,ren2020data,ma2023two,li2023state,li2023saliency,luo2023predicting}. Despite progress, two issues persist: (i) difficulty modeling both long-term global dependencies and fine-grained local patterns in degradation signals, and (ii) single-path architectures that are vulnerable to noise and information loss along the feature pipeline.

To address these gaps, we propose a Multiscale Dual-path Feature Aggregation Network (MDFA-Net) for LIB RUL forecasting. MDFA-Net separates representation learning into two complementary paths and then fuses them adaptively. The first path (MF-Net) emphasizes information preservation via dense connections and multiscale processing at the input to retain global trends while exposing multi-resolution cycles. The second path (EC-Net) couples CNN and Transformer blocks to jointly capture local context and content-based long-range interactions. A lightweight fusion module with position-enhanced attention assigns data-dependent weights to features from both paths, mitigating interference and highlighting degradation-relevant cues.

Our contributions are threefold:
\begin{itemize}
	\item We introduce MDFA-Net, a dual-path deep architecture tailored to LIB RUL forecasting that jointly captures global trends and local irregularities under noise and operating variations.
	\item We design MF-Net with dense connectivity and multiscale inputs for information-preserving, low-loss feature extraction, and EC-Net that integrates CNN (local) with Transformer (global) modeling for robust cross-scale dependencies.
	\item We develop a position-enhanced attention fusion that adaptively weights features from the two paths, improving robustness and focusing on degradation-informative components. Experiments on NASA and CALCE datasets demonstrate consistent gains over strong data-driven baselines in RUL estimation.
\end{itemize}
\section{Methodology}\label{sec:method}
We formulate RUL prediction for lithium-ion batteries (LIBs) and present MDFA-Net, a dual-path architecture that aggregates multi-scale local cues and global dependencies with a lightweight fusion head.

\subsection{MDFA-Net Overview}\label{subsec:overview}
\begin{figure}[t]
	\centering
	\includegraphics[scale=0.7]{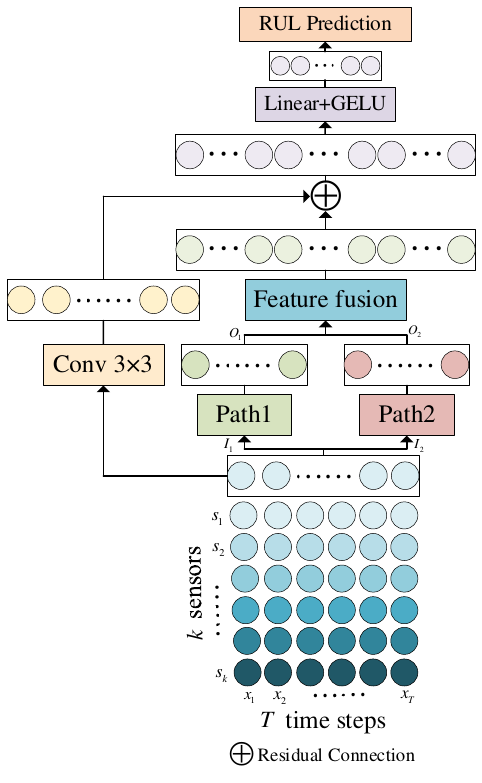}%
	\caption{Architecture of MDFA-Net.}
	\label{fig:mdfa}
\end{figure}

MDFA-Net consists of two paths (Fig.~\ref{fig:mdfa}): (i) a Multi-scale Feature path (MF-Net) built upon dense connectivity to preserve and reuse features under scarce data; (ii) an Encoding path (EC-Net) that couples self-attention (global interactions) with depthwise-separable convolution (local patterns). Path outputs are concatenated and projected:
\begin{equation}
	O_{\mathrm{fuse}}=\mathrm{Concat}(O^{(1)}, O^{(2)})\,W^{f}, \quad W^{f}\!\in\!\mathbb{R}^{2D_{\!model}\times D_{\!model}}.\label{eq:fuse}
\end{equation}
We inject relative positional information before self-attention to preserve temporal ordering. A final linear head yields the RUL prediction.

\subsubsection{Path I: Multi-scale Feature Network (MF-Net)}\label{subsec:mfnet}
\begin{figure}[t]
	\centering
	\includegraphics[scale=1.45]{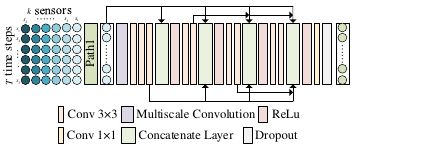}%
	\caption{The diagram of the proposed first path multiscale feature network (MF-Net).}
	\label{fig:mfnet}
\end{figure}

MF-Net (Fig.~\ref{fig:mfnet}) stacks a multi-scale stem with densely connected blocks to maintain stable gradients and high feature reuse. The stem applies parallel $1{\times}1$, $3{\times}3$, $5{\times}5$, $7{\times}7$ convolutions to capture short/medium/long temporal contexts, then compresses with a $1{\times}1$ projection \cite{ding2022scaling,li2021novel,chen2021time,hu2023fcan}. Dense concatenation across blocks preserves original and newly formed features with minimal overhead.

\subsubsection{Path II: Encoding Network (EC-Net)}\label{subsec:ecnet}
\begin{figure}[t]
	\centering
	\includegraphics[scale=0.56]{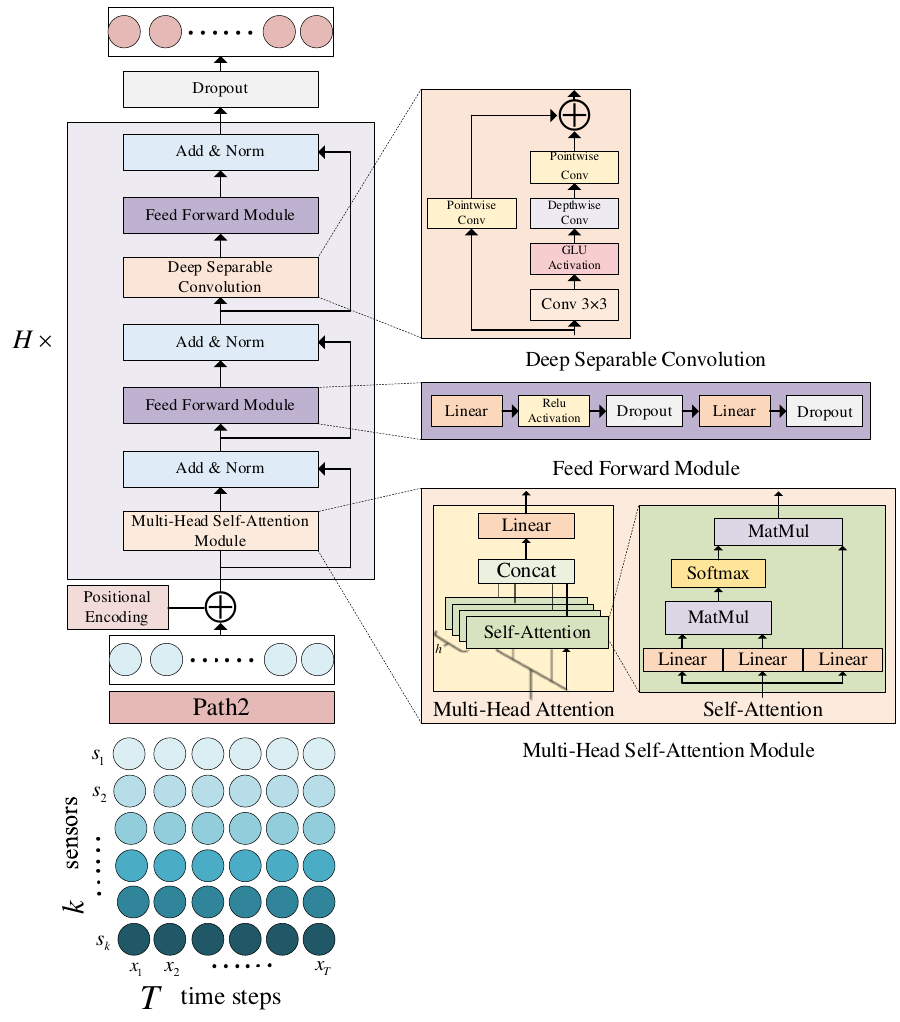}
	\caption{The diagram of the proposed second path encoding network (EC-Net).} 
	\label{fig:ecnet}
\end{figure}

EC-Net (Fig.~\ref{fig:ecnet}) adopts an Attention–FFN–Conv–FFN layout to first model long-range interactions, then enhance local selectivity via depthwise-separable convolution, followed by non-linear refinement:
\begin{align}
	\mathrm{MHSA}(Q,K,V) &= \mathrm{softmax}\!\left(\frac{QK^\top}{\sqrt{d_k}}\right)V,\label{eq:attn}\\
	\mathrm{FFN}(X) &= \sigma(XW_1{+}b_1)W_2{+}b_2.\label{eq:ffn}
\end{align}
This split ordering can be viewed as a Lie–Trotter style alternation between interaction (attention) and convection-like local transport (convolution), offering a practical balance between global coherence and local sharpness.

\section{Experiments}\label{sec:exp}
We evaluate MDFA-Net on two public LIB datasets and compare against recent RUL predictors. We also analyze design choices via ablations.

\subsection{Datasets}\label{subsec:datasets}
We use cyclic-test capacity datasets from NASA and CALCE, which exhibit diverse operating profiles and local variations \cite{dong2018battery,lin2021state,roman2021machine,li2020remaining}. Following common practice, the end-of-life (EOL) is set to \(70\%\) nominal capacity (1.40Ah for NASA; 0.77Ah for CALCE). We adopt a leave-one-out protocol: one cell is used for testing and the remaining cells for training.

\begin{table*}[t]
	\caption{Results on the NASA dataset (best baseline marked with *, best overall in \textbf{bold}).}
	\label{tab:nasa}
	\footnotesize
	\setlength{\tabcolsep}{4pt} 
	\centering
	\scalebox{0.55}{
		\begin{tabular}{lccccccccccccccc}
			\toprule
			\multirow{2}{*}{Method} &
			\multicolumn{3}{c}{B0005} & \multicolumn{3}{c}{B0006} & \multicolumn{3}{c}{B0007} & \multicolumn{3}{c}{B0018} & \multicolumn{3}{c}{Avg}\\
			\cmidrule(lr){2-4}\cmidrule(lr){5-7}\cmidrule(lr){8-10}\cmidrule(lr){11-13}\cmidrule(lr){14-16}
			& \(R^2\) & $E_{MAE}$ & $E_{RMSE}$ & \(R^2\) & $E_{MAE}$ & $E_{RMSE}$ & \(R^2\) & $E_{MAE}$ & $E_{RMSE}$ & \(R^2\) & $E_{MAE}$ & $E_{RMSE}$ & \(R^2\) & $E_{MAE}$ & $E_{RMSE}$ \\
			\midrule
			LSTM \cite{zhang2018long} & 0.799 & 0.0672 & 0.0802 & 0.6711 & 0.1123 & 0.1258 & 0.2369 & 0.1162 & 0.1301 & 0.8546 & 0.0420 & 0.0486 & 0.6404 & 0.0844 & 0.0962 \\
			DCNN \cite{li2018remaining} & 0.8967 & 0.0439 & 0.0575 & 0.8455 & 0.0687 & 0.0862 & 0.9205 & 0.0326 & 0.0420 & 0.7928 & 0.0424 & 0.0580 & 0.8639 & 0.0469 & 0.0609 \\
			CNN-LSTM \cite{kara2021data} & 0.9385 & 0.0313 & 0.0444 & 0.7005 & 0.1006 & 0.1200 & 0.7570 & 0.0587 & 0.0734 & 0.6968 & 0.0537 & 0.0701 & 0.7732 & 0.0611 & 0.0770 \\
			Lai et al. \cite{lai2022multi} & 0.9223 & 0.0453 & 0.0499 & 0.9196 & 0.0473 & 0.0621 & 0.6308 & 0.0720 & 0.0905 & 0.8375 & 0.0364 & 0.0513 & 0.8275 & 0.0503 & 0.0635 \\
			DeTransformer \cite{chen2022transformer} & 0.9385 & 0.0332 & 0.0444 & 0.9656$^{*}$ & 0.0311$^{*}$ & 0.0406$^{*}$ & 0.9538 & 0.0247 & 0.0320 & 0.8510 & 0.0410 & 0.0548 & 0.9182 & 0.0325 & 0.0429 \\
			MMMe \cite{:185329} & 0.9387$^{*}$ & 0.0335 & 0.0443 & 0.9531 & 0.0370 & 0.0474 & 0.9598$^{*}$ & 0.0249 & 0.0298 & 0.8643$^{*}$ & 0.0343$^{*}$ & 0.0469$^{*}$ & 0.9290$^{*}$ & 0.0324 & 0.0421 \\
			AUKF-GASVR \cite{xue2020remaining} & 0.8753 & \textbf{0.0148}$^{*}$ & \textbf{0.0230}$^{*}$ & 0.7743 & 0.0392 & 0.0510 & 0.9547 & \textbf{0.0091}$^{*}$ & \textbf{0.0134}$^{*}$ & 0.5614 & 0.0382 & 0.0547 & 0.7914 & \textbf{0.0253}$^{*}$ & 0.0355$^{*}$ \\
			JPO-CFNN \cite{ansari2023optimized} & -- & -- & 0.0258 & -- & -- & 0.0462 & -- & -- & 0.0350 & -- & -- & 0.1851 & -- & -- & 0.0730 \\
			\textbf{MDFA-Net} & \textbf{0.9613} & 0.0277 & 0.0352 & \textbf{0.9709} & \textbf{0.0284} & \textbf{0.0374} & \textbf{0.9689} & 0.0205 & 0.0262 & \textbf{0.9274} & \textbf{0.0270} & \textbf{0.0343} & \textbf{0.9571} & 0.0259 & \textbf{0.0333} \\
			\bottomrule
	\end{tabular}}
	\vspace{-6pt}
\end{table*}

\subsection{Experimental Setup}\label{subsec:setup}

We use Bayesian search for model sizing. Window \(T_w{=}16\) (NASA) and \(T_w{=}64\) (CALCE) unless stated, Adam (MSE loss), lr \(0.01\), dropout 0.3, early stopping within 1000 epochs. All runs on a Windows 11 workstation (RTX 3080, 16GB RAM).

\textbf{Baselines.}
We compare with: (1) RNN/CNN-only \cite{zhang2018long,li2018remaining};
(2) enhanced RNN/CNN \cite{kara2021data,lai2022multi};
(3) hybrid/aging-aware models \cite{chen2022transformer,:185329,xue2020remaining,ansari2023optimized}.
For reproducible methods we average 10 runs, otherwise we cite reported bests.

\subsection{Results on NASA}\label{subsec:nasa}
\begin{figure}[t]
	\centering
	\subfloat[B0005]{\includegraphics[width=0.25\linewidth]{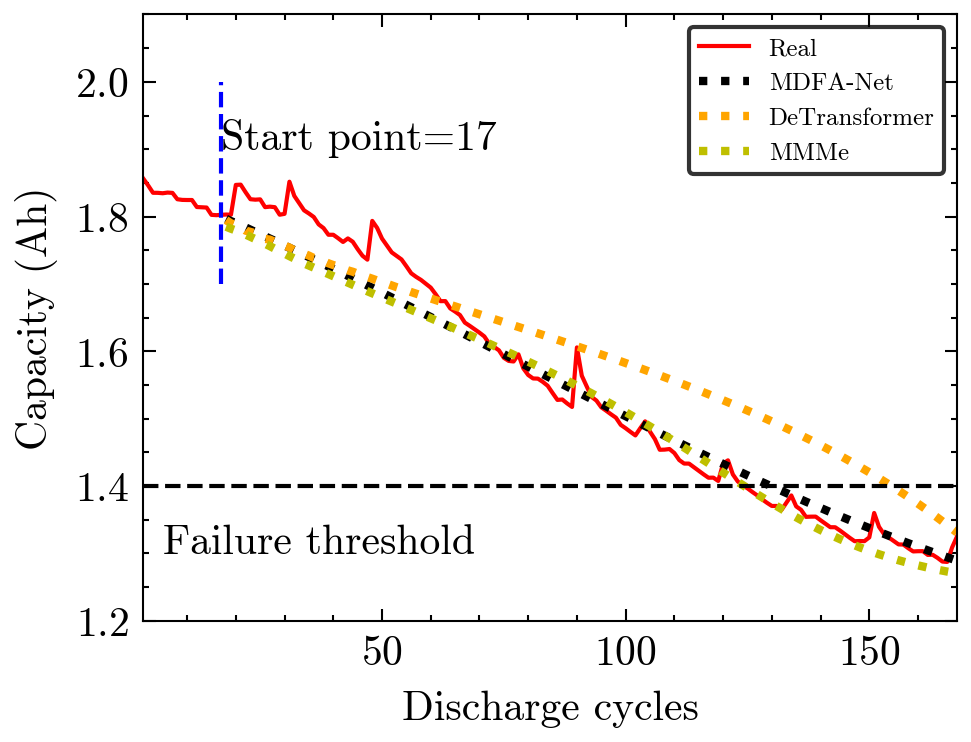}}\hfill
	\subfloat[B0006]{\includegraphics[width=0.25\linewidth]{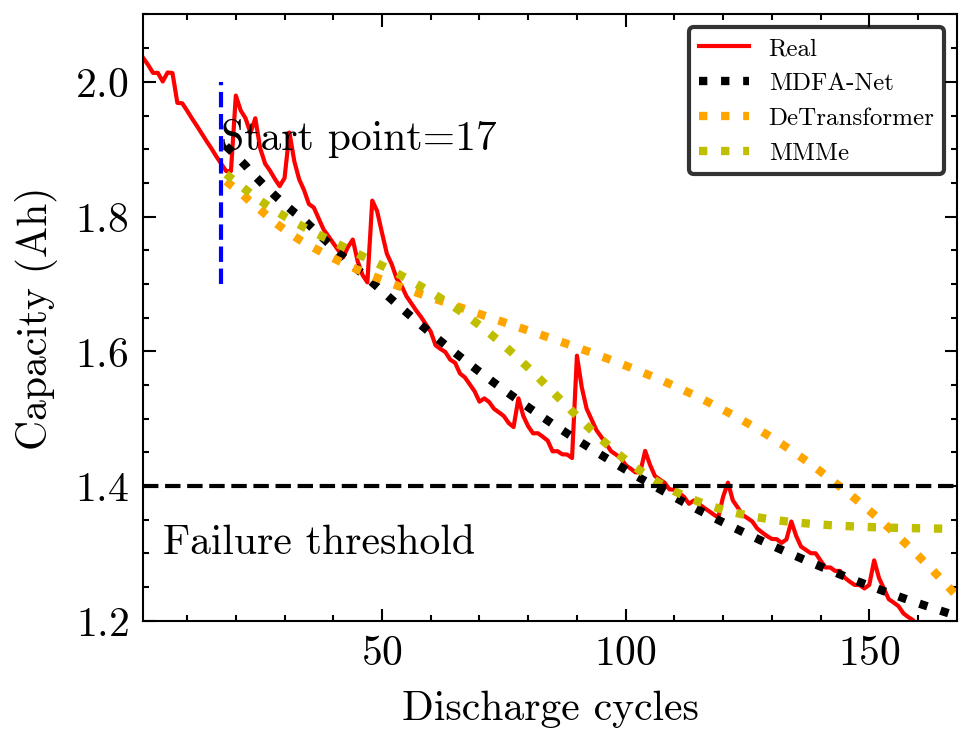}}\hfill
	\subfloat[B0007]{\includegraphics[width=0.25\linewidth]{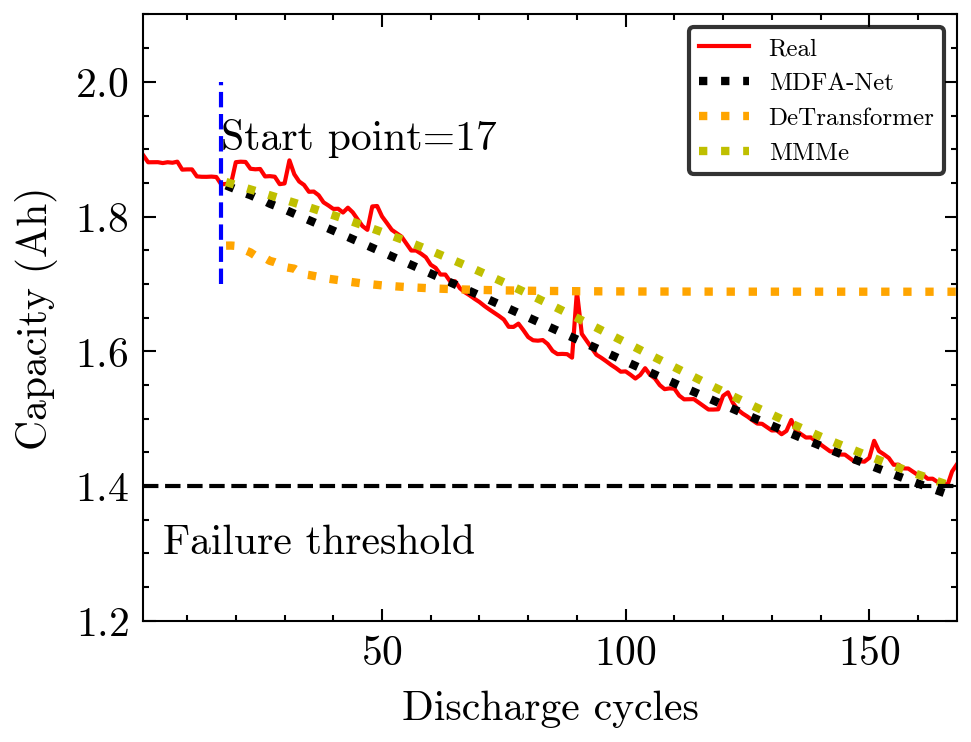}}\hfill
	\subfloat[B0018]{\includegraphics[width=0.25\linewidth]{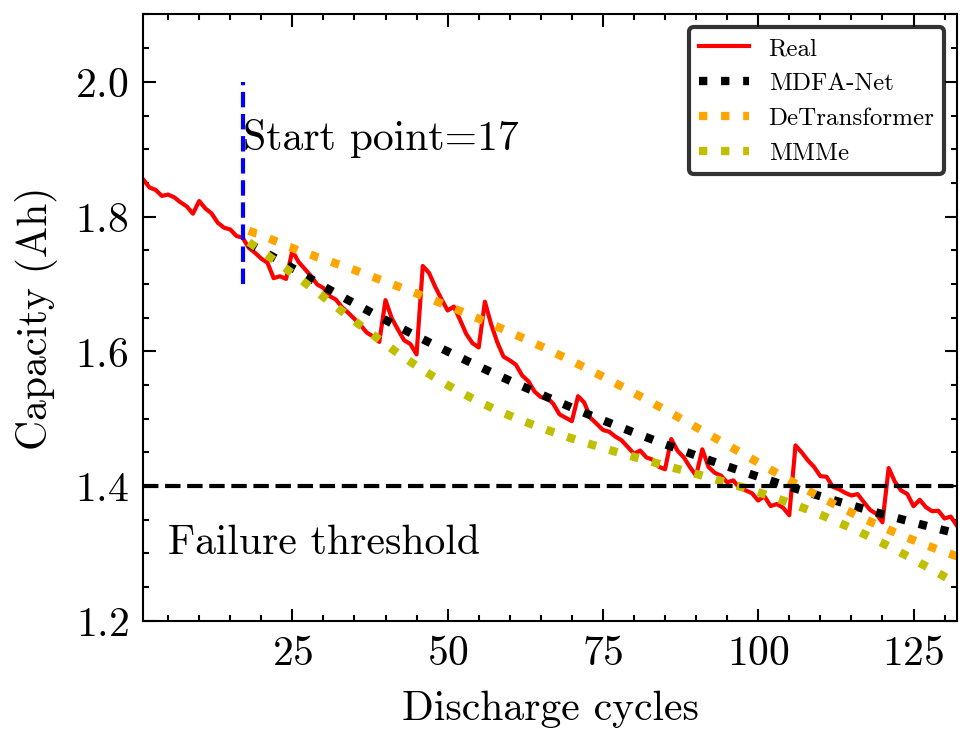}}
	\vspace{-6pt}
	\caption{RUL prediction results on four test LIB.}
	\label{fig:pred_nasa}
	\vspace{-8pt}
\end{figure}

Initially, our approach is evaluated against the established baseline utilizing the NASA dataset. Results of the final baseline predictions can be found in Table~\ref{tab:nasa}. Clearly, across all three performance metrics evaluated, our approach outshines the most recent baseline model significantly. This underscores the superior accuracy in RUL forecasting achieved by MDFA-Net. Crucially, MDFA-Net's enhancements compared to the top-performing current techniques are more pronounced for the B0006 and B0018 datasets. In particular, MDFA-Net achieves a 8.68\% reduction in $E_{MAE}$ for B0006, and for B0018, it enhances \(R^2\) by 7.30\% and lowers $E_{RMSE}$ by 26.86\% relative to existing benchmarks. Furthermore, it's noteworthy that MDEA-Net's enhancement is particularly significant with respect to \(R^2\), indicating a superior fit to the target values by our model. Qualitatively, Fig.~\ref{fig:pred_nasa} shows predicted vs.\ true RUL for a random cell from each NASA subset (also comparing DeTransformer and MMMe). MDFA-Net tracks target curves closely and tends to under-estimate rather than over-estimate RUL, which is safer in practice.

In practical Battery Management System(BMS) deployments, the efficiency-critical metric is online inference latency, not offline training time. Using B0006 as an example and keeping the setup identical to \cite{chen2022transformer} and MMMe \cite{:185329}, we report both training and inference time and explicitly acknowledge an efficiency–accuracy trade-off (see Table~\ref{tab:time}). MDFA-Net’s total test time is 0.7812 s for 151 steps, corresponding to ~5.17 ms per step, DeTransformer is 0.2343 s (1.55 ms per step), and MMMe is 1.2656 s (8.38 ms per step). Thus, MDFA-Net is slower than DeTransformer at inference, but it remains within the millisecond range suitable for real-time prediction on the inference side, while delivering higher accuracy.

\begin{table}[t]
	\caption{Training and inference time on B0006.}
	\label{tab:time}
	\centering
	\scalebox{1}{
		\begin{tabular}{lcccc}
			\toprule
			Method & Train(s) & Test(s) & \(R^2\) & $E_{RMSE}$\\
			\midrule
			MDFA-Net & 243.18 & 0.7812 & 0.9709 & 0.0374\\
			DeTransformer \cite{chen2022transformer} & 154.64 & 0.2343 & 0.9656 & 0.0406\\
			MMMe \cite{:185329} & 491.35 & 1.2656 & 0.9531 & 0.0474\\
			\bottomrule
	\end{tabular}}
	\vspace{-6pt}
\end{table}

\subsection{Results on CALCE}\label{subsec:calce}
\begin{figure}[t]
	\centering
	\subfloat[CS2\_35]{\includegraphics[width=0.25\linewidth]{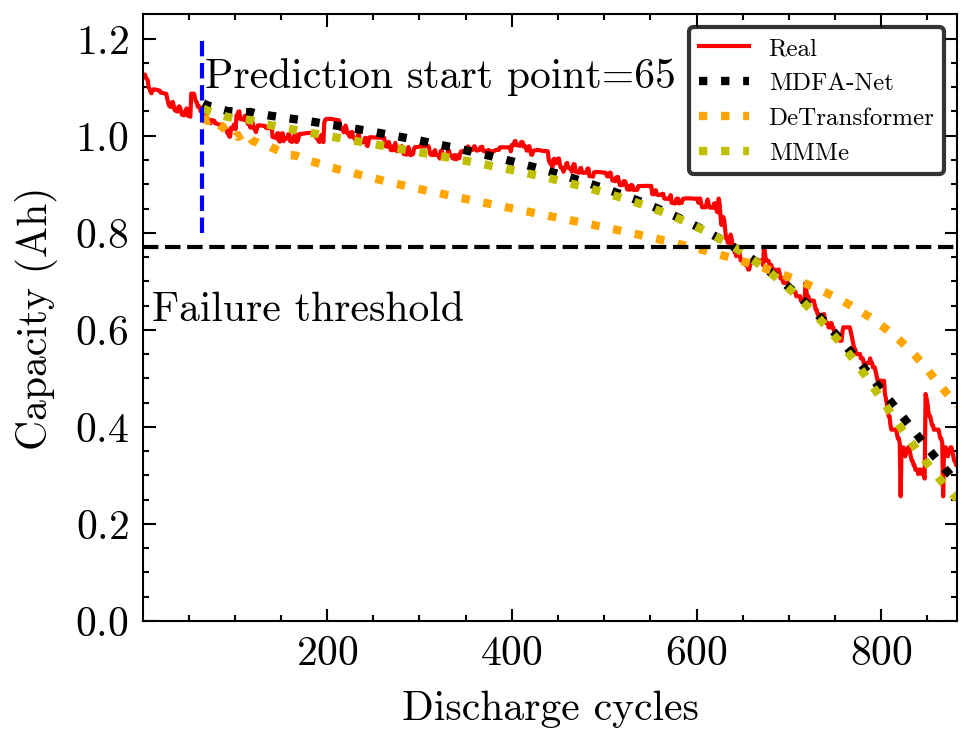}}\hfill
	\subfloat[CS2\_36]{\includegraphics[width=0.25\linewidth]{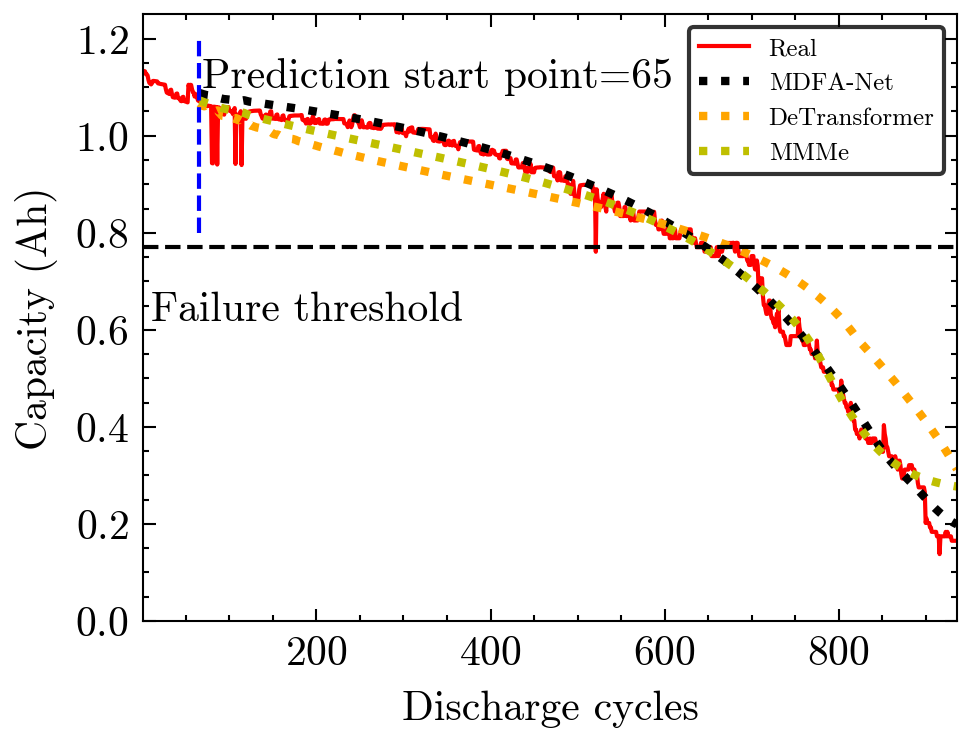}}\hfill
	\subfloat[CS2\_37]{\includegraphics[width=0.25\linewidth]{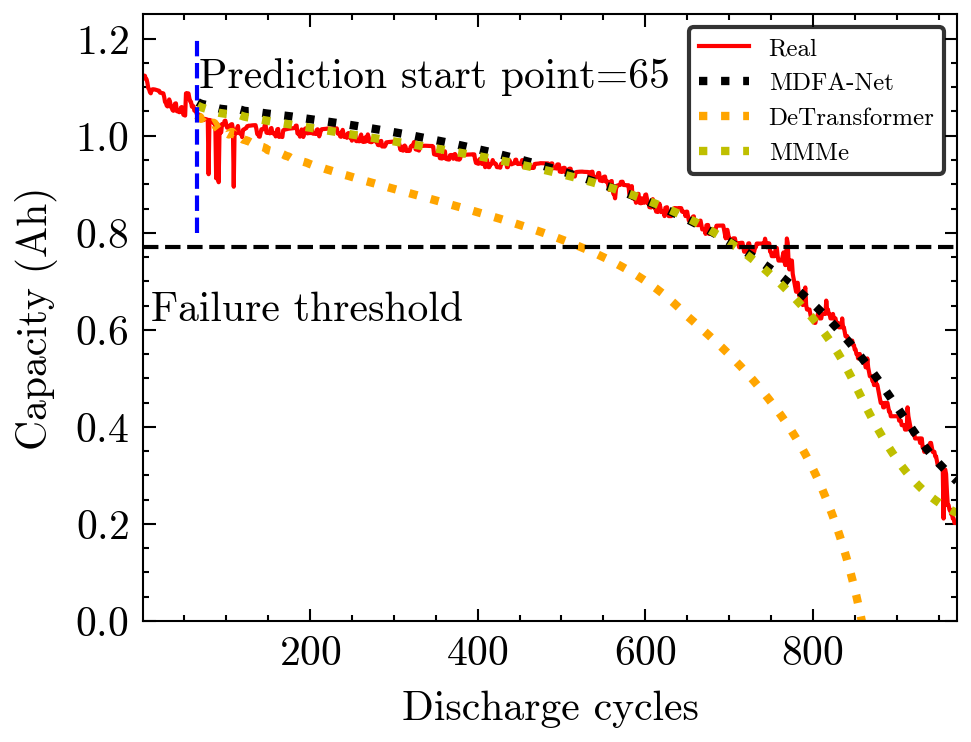}}\hfill
	\subfloat[CS2\_38]{\includegraphics[width=0.25\linewidth]{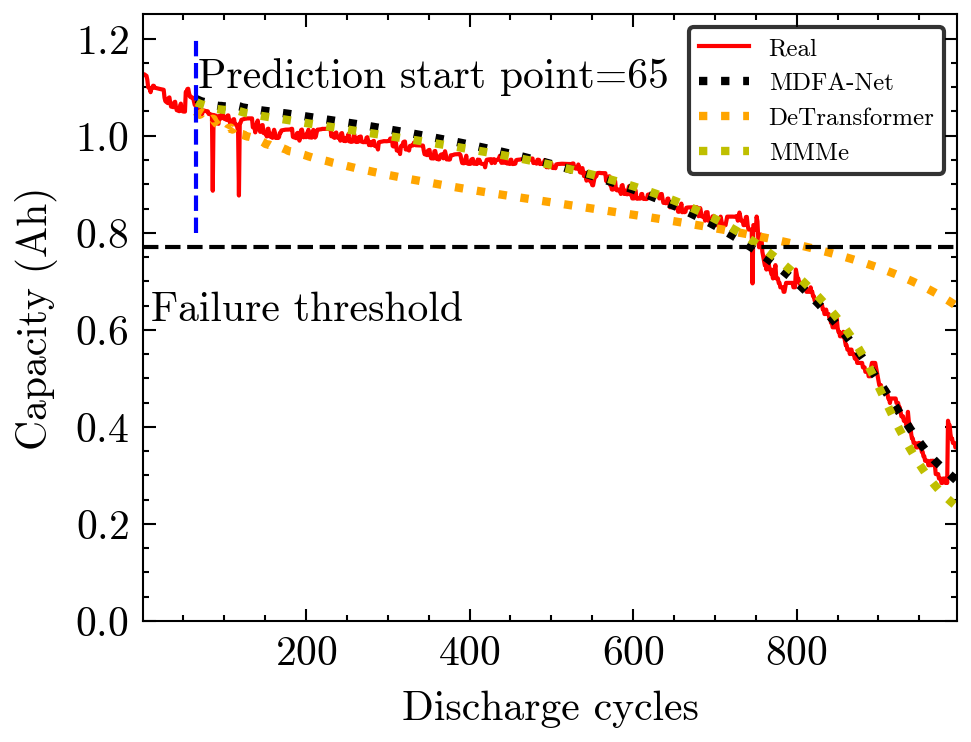}}
	\vspace{-6pt}
	\caption{RUL prediction results on CALCE.}
	\label{fig:pred_calce}
	\vspace{-8pt}
\end{figure}

In this experiment, the identical MDFA-Net model from the NASA study was utilized, featuring a sliding time window size of $T_{w}=64 $. Table~\ref{tab:calce} outlines the comparative results between MDFA-Net and various leading-edge approaches. It is evident that MDFA-Net surpasses all baseline models by a significant margin. Relative to the top-performing contemporary techniques,  the MDFA-Net we propose shows enhancements of 0.47\%, 5.67\%, and 9.47\% in \(R^2\), $E_{MAE}$, and $E_{RMSE}$ metrics, respectively. Furthermore, Fig.~\ref{fig:pred_calce}'s juxtaposition of estimated versus genuine battery RUL underscores MDFA-Net's performance on this dataset.

\begin{table}[t]
	\caption{Results on the CALCE dataset.}
	\label{tab:calce}
	\centering
	\scalebox{1}{
		\begin{tabular}{lccc}
			\toprule
			Method & \(R^2\) & $E_{MAE}$ & $E_{RMSE}$\\
			\midrule
			LSTM \cite{zhang2018long} & 0.9486 & 0.0401 & 0.0478 \\
			DCNN \cite{li2018remaining} & 0.9613 & 0.0319 & 0.0427 \\
			CNN-LSTM \cite{kara2021data} & 0.7482 & 0.0915 & 0.1074 \\
			Lai et al. \cite{lai2022multi} & 0.5971 & 0.1028 & 0.1227 \\
			DeTransformer \cite{chen2022transformer} & 0.9782$^{*}$ & 0.0245 & 0.0319 \\
			MMMe \cite{:185329} & 0.9763 & 0.0229$^{*}$ & 0.0306$^{*}$ \\
			\textbf{MDFA-Net} & \textbf{0.9828} & \textbf{0.0216} & \textbf{0.0277} \\
			\bottomrule
	\end{tabular}}
	\vspace{-6pt}
\end{table}

\subsection{Analysis of MDFA-Net}\label{subsec:analysis}
\subsubsection{Impact of window size}\label{subsec:window}
\begin{figure}[t]
	\centering
	\includegraphics[width=0.33\linewidth]{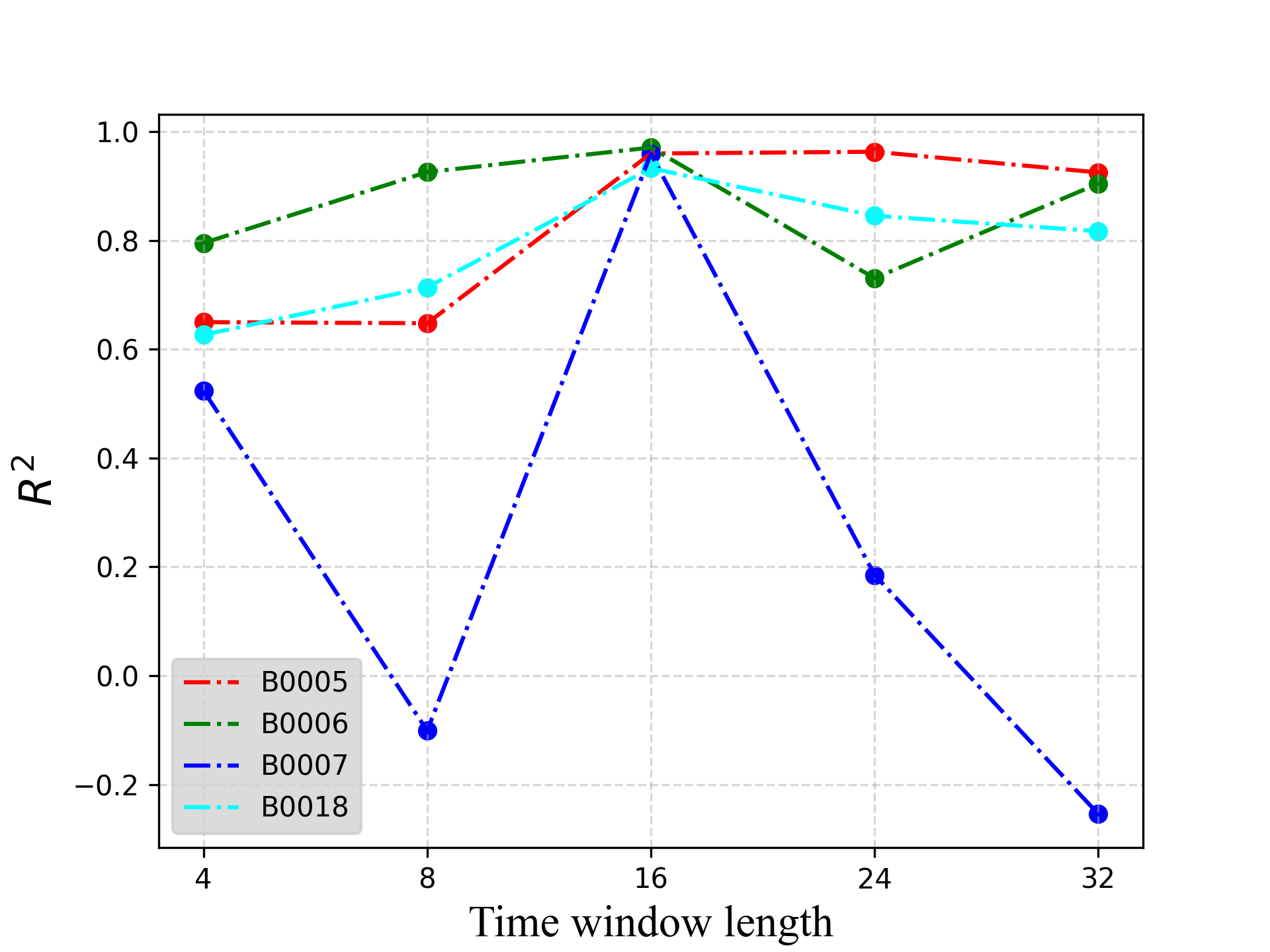}\hfill
	\includegraphics[width=0.33\linewidth]{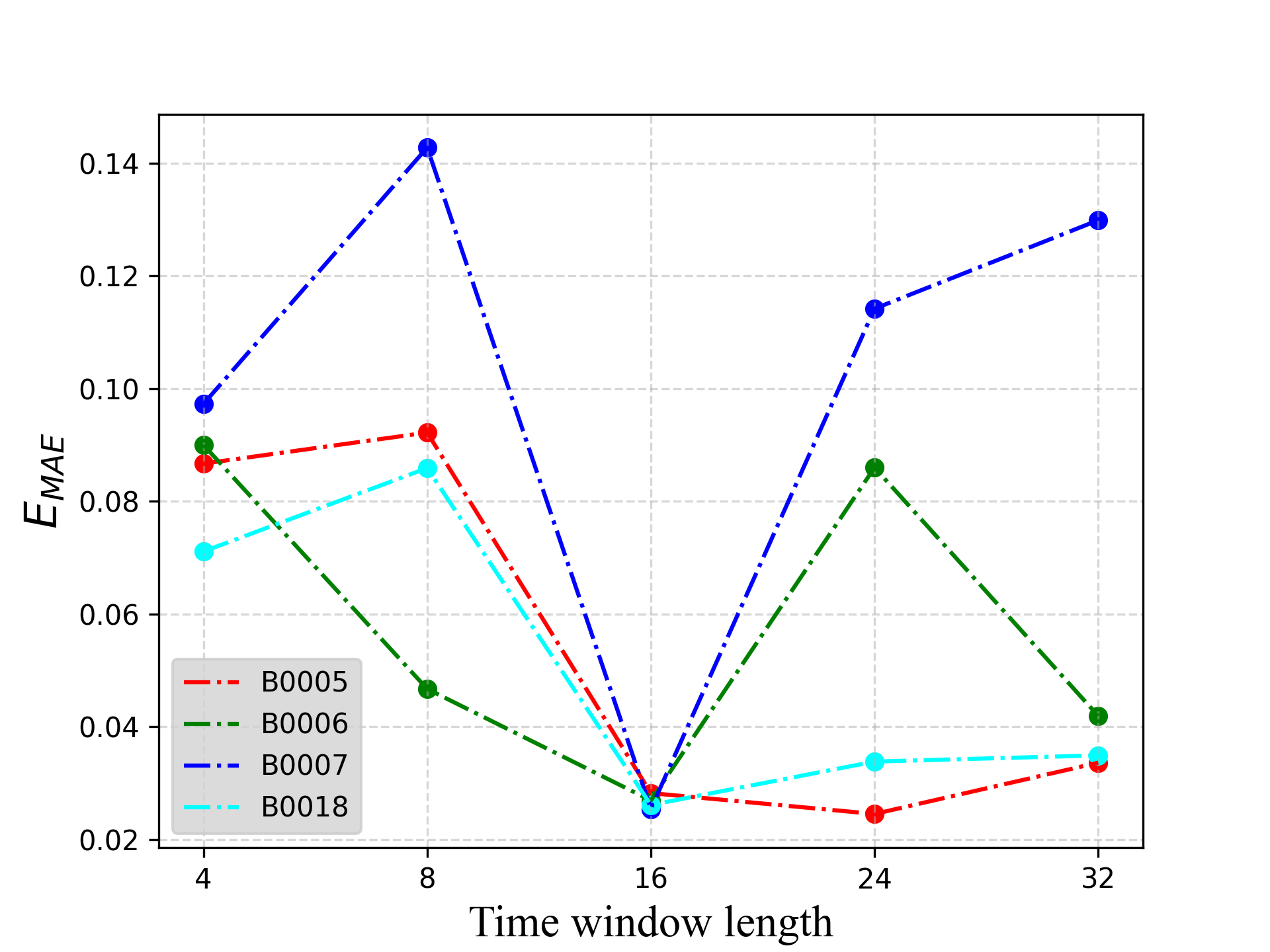}\hfill
	\includegraphics[width=0.33\linewidth]{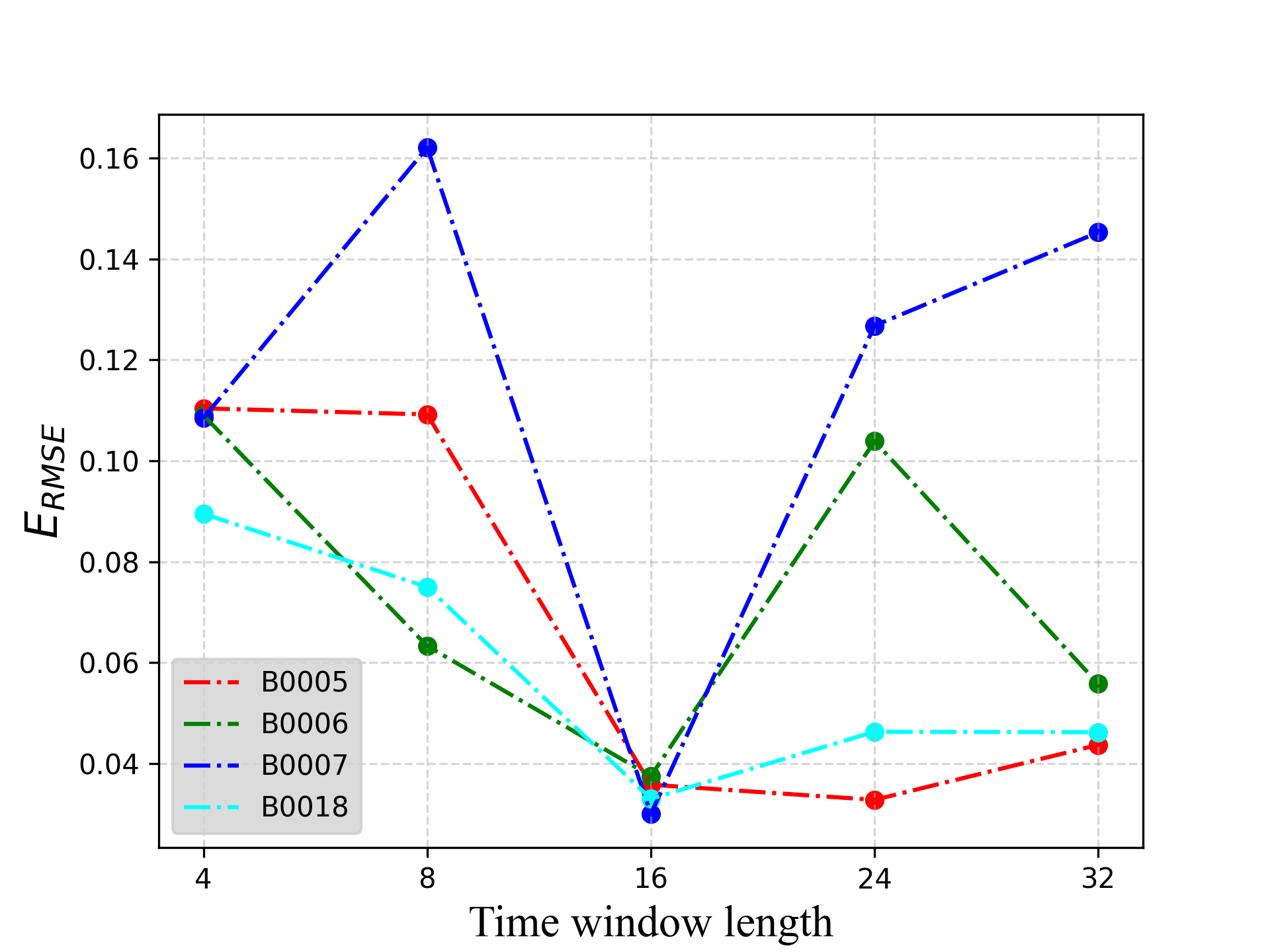}
	\caption{Performance of MDFA-Net with different time window sizes on the NASA dataset.}
	\label{fig:window}
\end{figure}

We sweep \(T_w\in\{4,8,16,24,32\}\) on NASA. As shown in Fig.~\ref{fig:window}, \(T_w{=}16\) yields the best trade-off on B0006, B0007 and B0018, whereas B0005 benefits from \(T_w{=}24\) due to more complex operating conditions; \(T_w{=}32\) over-smooths short-term variations.

\subsubsection{Ablation study}\label{subsec:ablation}
We examine the roles of (i) positional encoding (PE) and self-attention in fusion, and (ii) the dual-path design (parallel vs. series). Results on NASA and CALCE (Table~\ref{tab:ablation}) also reveal an imbalance between the two branches: MF-Net alone provides a strong baseline, while adding EC-Net yields consistent but limited average gains. This aligns with the data characteristics. Degradation in NASA and CALCE is predominantly smooth and monotonic with few local bursts; therefore the information-preserving multiscale local branch (MF-Net) explains most of the variance. EC-Net is not redundant but complementary: it supplies content-based long-term cues and local corrections that the fusion module exploits when beneficial.
\begin{table}[t]
	\caption{Ablations on NASA and CALCE.}
	\label{tab:ablation}
	\centering
	\scalebox{0.75}{
		\begin{tabular}{lcccccc}
			\toprule
			\multirow{2}{*}{Method} & \multicolumn{3}{c}{NASA} & \multicolumn{3}{c}{CALCE} \\
			\cmidrule(lr){2-4}\cmidrule(lr){5-7}
			& \(R^2\) & $E_{MAE}$ & $E_{RMSE}$ & \(R^2\) & $E_{MAE}$ &$E_{RMSE}$ \\
			\midrule
			MDFA-Net w/o positional encoding & 0.9559 & 0.0268 & 0.0339 & 0.9784 & 0.0251 & 0.0319 \\
			MDFA-Net w/o PE \& Self-Attn & 0.9560 & 0.0272 & 0.0335 & 0.9807 & 0.0223 & 0.0293 \\
			MDFA-Net w/o MF-Net & 0.9133 & 0.0372 & 0.0456 & 0.9448 & 0.0403 & 0.0493 \\
			MDFA-Net w/o EC-Net & 0.9537 & 0.0274 & 0.0355 & 0.9811 & 0.0218 & 0.0295 \\
			MDFA-Net in series & 0.9275 & 0.0361 & 0.0450 & 0.9819 & 0.0218 & 0.0285 \\
			MDFA-Net & \textbf{0.9571} & \textbf{0.0259} & \textbf{0.0333} & \textbf{0.9828} & \textbf{0.0216} & \textbf{0.0277} \\
			\bottomrule
	\end{tabular}}
	\vspace{-8pt}
\end{table}

\section{Conclusions and future work}\label{sec4} 

We propose MDFA-Net, a multiscale dual-path feature aggregation model for RUL prediction. Two parallel paths extract complementary deep and shallow features to capture local–global dependencies, while a self-attention fusion module with positional cues adaptively weights the paths. Ablation studies validate the design, and tests on two LIB degradation datasets show consistent gains over recent deep-learning baselines. Next, we will employ transfer learning to address limited training data.


\bibliographystyle{unsrt} 
\bibliography{ref} 

@article{lv2024remaining,
	title={Remaining useful life prediction for lithium-ion batteries incorporating spatio-temporal information},
	author={Lv, Zihao and Song, Yi and He, Chunlin and Xu, Liming},
	journal={Journal of Energy Storage},
	volume={88},
	pages={111626},
	year={2024},
	publisher={Elsevier}
}

@article{li2025learning,
	title={Learning error refinement in stochastic gradient descent-based latent factor analysis via diversified pid controllers},
	author={Li, Jinli and Yuan, Ye and Luo, Xin},
	journal={IEEE Transactions on Emerging Topics in Computational Intelligence},
	year={2025},
	publisher={IEEE}
}

@article{wang2024gt,
	title={GT-A 2 T: Graph tensor alliance attention network},
	author={Wang, Ling and Liu, Kechen and Yuan, Ye},
	journal={IEEE/CAA Journal of Automatica Sinica},
	year={2024},
	publisher={IEEE}
}

@article{chen2024sdgnn,
	title={Sdgnn: Symmetry-preserving dual-stream graph neural networks},
	author={Chen, Jiufang and Yuan, Ye and Luo, Xin},
	journal={IEEE/CAA journal of automatica sinica},
	volume={11},
	number={7},
	pages={1717--1719},
	year={2024},
	publisher={IEEE}
}

@article{luo2024pseudo,
	title={Pseudo gradient-adjusted particle swarm optimization for accurate adaptive latent factor analysis},
	author={Luo, Xin and Chen, Jiufang and Yuan, Ye and Wang, Zidong},
	journal={IEEE Transactions on Systems, Man, and Cybernetics: Systems},
	volume={54},
	number={4},
	pages={2213--2226},
	year={2024},
	publisher={IEEE}
}

@article{li2023nonlinear,
	title={A nonlinear PID-incorporated adaptive stochastic gradient descent algorithm for latent factor analysis},
	author={Li, Jinli and Luo, Xin and Yuan, Ye and Gao, Shangce},
	journal={IEEE Transactions on Automation Science and Engineering},
	volume={21},
	number={3},
	pages={3742--3756},
	year={2023},
	publisher={IEEE}
}

@article{chen2024state,
	title={A state-migration particle swarm optimizer for adaptive latent factor analysis of high-dimensional and incomplete data},
	author={Chen, Jiufang and Liu, Kechen and Luo, Xin and Yuan, Ye and Sedraoui, Khaled and Al-Turki, Yusuf and Zhou, MengChu},
	journal={IEEE/CAA Journal of Automatica Sinica},
	volume={11},
	number={11},
	pages={2220--2235},
	year={2024},
	publisher={IEEE}
}

@article{yuan2022kalman,
	title={A Kalman-filter-incorporated latent factor analysis model for temporally dynamic sparse data},
	author={Yuan, Ye and Luo, Xin and Shang, Mingsheng and Wang, Zidong},
	journal={IEEE Transactions on Cybernetics},
	volume={53},
	number={9},
	pages={5788--5801},
	year={2022},
	publisher={IEEE}
}

@article{yuan2023adaptive,
	title={An adaptive divergence-based non-negative latent factor model},
	author={Yuan, Ye and Wang, Renfang and Yuan, Guangxiao and Xin, Luo},
	journal={IEEE Transactions on Systems, Man, and Cybernetics: Systems},
	volume={53},
	number={10},
	pages={6475--6487},
	year={2023},
	publisher={IEEE}
}

@article{yuan2020multilayered,
	title={A multilayered-and-randomized latent factor model for high-dimensional and sparse matrices},
	author={Yuan, Ye and He, Qiang and Luo, Xin and Shang, Mingsheng},
	journal={IEEE transactions on big data},
	volume={8},
	number={3},
	pages={784--794},
	year={2020},
	publisher={IEEE}
}

@article{Wang2024,
	author = {Wang, Hongjun and Song, Yi and Chen, Wei and Luo, Zhipeng and Li, Chongshou and Li, Tianrui},
	title = {A Survey of Co-Clustering},
	year = {2024},
	issue_date = {November 2024},
	publisher = {Association for Computing Machinery},
	address = {New York, NY, USA},
	volume = {18},
	number = {9},
	issn = {1556-4681}}

@article{luo2020position,
	title={Position-transitional particle swarm optimization-incorporated latent factor analysis},
	author={Luo, Xin and Yuan, Ye and Chen, Sili and Zeng, Nianyin and Wang, Zidong},
	journal={IEEE Transactions on Knowledge and Data Engineering},
	volume={34},
	number={8},
	pages={3958--3970},
	year={2020},
	publisher={IEEE}
}

@article{shang2021alpha,
	title={An $\alpha$--$\beta$-divergence-generalized recommender for highly accurate predictions of missing user preferences},
	author={Shang, Mingsheng and Yuan, Ye and Luo, Xin and Zhou, MengChu},
	journal={IEEE transactions on cybernetics},
	volume={52},
	number={8},
	pages={8006--8018},
	year={2021},
	publisher={IEEE}
}

@article{xin2019non,
	title={Non-negative latent factor model based on $\beta$-divergence for recommender systems},
	author={Xin, Luo and Yuan, Ye and Zhou, MengChu and Liu, Zhigang and Shang, Mingsheng},
	journal={IEEE Transactions on Systems, Man, and Cybernetics: Systems},
	volume={51},
	number={8},
	pages={4612--4623},
	year={2019},
	publisher={IEEE}
}

@inproceedings{yuan2020generalized,
	title={A generalized and fast-converging non-negative latent factor model for predicting user preferences in recommender systems},
	author={Yuan, Ye and Luo, Xin and Shang, Mingsheng and Wu, Di},
	booktitle={Proceedings of The Web Conference 2020},
	pages={498--507},
	year={2020}
}

@article{lv2025state,
	title={State estimation of Lithium-ion Batteries with state space model},
	author={Lv, Zihao and Song, Yi and Xue, Yu and Xu, Shijie and He, Chunlin and Xu, Liming},
	journal={Engineering Applications of Artificial Intelligence},
	volume={159},
	pages={111463},
	year={2025},
	publisher={Elsevier}
}

@inproceedings{han2025sgd,
	title={Sgd-dyg: Self-reliant global dependency apprehending on dynamic graphs},
	author={Han, Minglian and Wang, Ling and Yuan, Ye and Luo, Xin},
	booktitle={Proceedings of the 31st ACM SIGKDD Conference on Knowledge Discovery and Data Mining V. 2},
	pages={802--813},
	year={2025}
}

@incollection{yuan2020temporal,
	title={Temporal web service qos prediction via kalman filter-incorporated latent factor analysis},
	author={Yuan, Ye and Shang, Mingsheng and Luo, Xin},
	booktitle={ECAI 2020},
	pages={561--568},
	year={2020},
	publisher={IOS Press}
}

@article{chen2021hyper,
	title={Hyper-parameter-evolutionary latent factor analysis for high-dimensional and sparse data from recommender systems},
	author={Chen, Jiufang and Yuan, Ye and Ruan, Tao and Chen, Jia and Luo, Xin},
	journal={Neurocomputing},
	volume={421},
	pages={316--328},
	year={2021},
	publisher={Elsevier}
}

@article{li2021proportional,
	title={A proportional-integral-derivative-incorporated stochastic gradient descent-based latent factor analysis model},
	author={Li, Jinli and Yuan, Ye and Ruan, Tao and Chen, Jia and Luo, Xin},
	journal={Neurocomputing},
	volume={427},
	pages={29--39},
	year={2021},
	publisher={Elsevier}
}

@article{yuan2018effects,
	title={Effects of preprocessing and training biases in latent factor models for recommender systems},
	author={Yuan, Ye and Luo, Xin and Shang, Ming-Sheng},
	journal={Neurocomputing},
	volume={275},
	pages={2019--2030},
	year={2018},
	publisher={Elsevier}
}

@article{he2024structure,
	title={Structure-preserved self-attention for fusion image information in multiple color spaces},
	author={He, Zhu and Lin, Mingwei and Luo, Xin and Xu, Zeshui},
	journal={IEEE Transactions on Neural Networks and Learning Systems},
	year={2024},
	publisher={IEEE}
}

@article{wu2024outlier,
	title={An outlier-resilient autoencoder for representing high-dimensional and incomplete data},
	author={Wu, Di and Hu, Yuanpeng and Liu, Kechen and Li, Jing and Wang, Xianmin and Deng, Song and Zheng, Nenggan and Luo, Xin},
	journal={IEEE Transactions on Emerging Topics in Computational Intelligence},
	year={2024},
	publisher={IEEE}
}

@article{wu2024online,
	title={Online sparse streaming feature selection with gaussian copula'},
	author={Wu, D and Li, Z and Chen, F and He, J and Luo, X},
	journal={IEEE Trans. Big Data},
	volume={10},
	number={1},
	year={2024}
}

@article{tang2025neural,
	title={Neural tucker factorization},
	author={Tang, Peng and Luo, Xin},
	journal={IEEE/CAA Journal of Automatica Sinica},
	volume={12},
	number={2},
	pages={475--477},
	year={2025},
	publisher={IEEE}
}

@article{chen2024latent,
	title={Latent-factorization-of-tensors-incorporated battery cycle life prediction},
	author={Chen, Minzhi and Tao, Li and Lou, Jungang and Luo, Xin},
	journal={IEEE/CAA Journal of Automatica Sinica},
	year={2024},
	publisher={IEEE}
}

@article{wu2023robust,
	title={Robust low-rank latent feature analysis for spatiotemporal signal recovery},
	author={Wu, Di and Li, Zechao and Yu, Zhikai and He, Yi and Luo, Xin},
	journal={IEEE Transactions on Neural Networks and Learning Systems},
	year={2023},
	publisher={IEEE}
}

@article{liu2020remaining,
	title={Remaining useful life prediction using a novel feature-attention-based end-to-end approach},
	author={Liu, Hui and Liu, Zhenyu and Jia, Weiqiang and Lin, Xianke},
	journal={IEEE Transactions on Industrial Informatics},
	volume={17},
	number={2},
	pages={1197--1207},
	year={2020},
	publisher={IEEE}
}

@article{liu2020cnn,
	title={CNN-FCM: System modeling promotes stability of deep learning in time series prediction},
	author={Liu, Penghui and Liu, Jing and Wu, Kai},
	journal={Knowledge-Based Systems},
	volume={203},
	pages={106081},
	year={2020},
	publisher={Elsevier}
}

@article{chen2024robust,
	title={A robust and efficient ensemble of diversified evolutionary computing algorithms for accurate robot calibration},
	author={Chen, Tinghui and Li, Shuai and Qiao, Yan and Luo, Xin},
	journal={IEEE Transactions on Instrumentation and Measurement},
	volume={73},
	pages={1--14},
	year={2024},
	publisher={IEEE}
}

@article{zeng2024novel,
	title={A novel tensor decomposition-based efficient detector for low-altitude aerial objects with knowledge distillation scheme},
	author={Zeng, Nianyin and Li, Xinyu and Wu, Peishu and Li, Han and Luo, Xin},
	journal={IEEE/CAA Journal of Automatica Sinica},
	volume={11},
	number={2},
	pages={487--501},
	year={2024},
	publisher={IEEE}
}

@inproceedings{lv2023study,
	title={A study of Chinese medicine entity recognition method by fusing multi-features and pointer networks},
	author={Lv, Zihao and He, Chunlin and Xu, Liming},
	booktitle={2023 IEEE International Conference on Systems, Man, and Cybernetics (SMC)},
	pages={2087--2092},
	year={2023},
	organization={IEEE}
}

@article{li2025semi,
	title={Semi-supervised structured nonnegative matrix factorization for anchor graph embedding},
	author={Li, Xiangli and Mei, Jianping and Mo, Yuanjian},
	journal={Neurocomputing},
	pages={130222},
	year={2025},
	publisher={Elsevier}
}

@article{mei2024dual,
	title={Dual semi-supervised hypergraph regular multi-view NMF with anchor graph embedding},
	author={Mei, Jianping and Li, Xiangli and Mo, Yuanjian},
	journal={Knowledge-Based Systems},
	volume={305},
	pages={112662},
	year={2024},
	publisher={Elsevier}
}

@article{WANG2025121315,
	title = {MCKP: Multi-aspect contextual knowledge-enhanced prompting for conversational recommender systems},
	journal = {Information Sciences},
	volume = {686},
	pages = {121315},
	year = {2025},
	issn = {0020-0255}
}

@article{WANG2023111129,
	title = {Enhancing conversational recommender systems via multi-level knowledge modeling with semantic relations},
	journal = {Knowledge-Based Systems},
	volume = {282},
	pages = {111129},
	year = {2023},
	issn = {0950-7051}
}

@article{mo2024semi,
	title={Semi-supervised nonnegative matrix factorization with label propagation and constraint propagation},
	author={Mo, Yuanjian and Li, Xiangli and Mei, Jianping},
	journal={Engineering Applications of Artificial Intelligence},
	volume={133},
	pages={108196},
	year={2024},
	publisher={Elsevier}
}

@article{luo2021fast,
	title={Fast and accurate non-negative latent factor analysis of high-dimensional and sparse matrices in recommender systems},
	author={Luo, Xin and Zhou, Yue and Liu, Zhigang and Zhou, MengChu},
	journal={IEEE Transactions on Knowledge and Data Engineering},
	volume={35},
	number={4},
	pages={3897--3911},
	year={2021},
	publisher={IEEE}
}

@article{wu2023graph,
	title={A graph-incorporated latent factor analysis model for high-dimensional and sparse data},
	author={Wu, Di and He, Yi and Luo, Xin},
	journal={IEEE transactions on emerging topics in computing},
	volume={11},
	number={4},
	pages={907--917},
	year={2023},
	publisher={IEEE}
}

@article{chen2023tensor,
	title={Tensor distribution regression based on the 3D conventional neural networks},
	author={Chen, Lin and Luo, Xin},
	journal={IEEE/CAA Journal of Automatica Sinica},
	volume={10},
	number={7},
	pages={1628--1630},
	year={2023},
	publisher={IEEE}
}

@article{yuan2024node,
	title={A node-collaboration-informed graph convolutional network for highly accurate representation to undirected weighted graph},
	author={Yuan, Ye and Wang, Ying and Luo, Xin},
	journal={IEEE Transactions on Neural Networks and Learning Systems},
	year={2024},
	publisher={IEEE}
}

@article{lin20243d,
	title={A 3d convolution-incorporated dimension preserved decomposition model for traffic data prediction},
	author={Lin, Mingwei and Liu, Jiaqi and Chen, Hong and Xu, Xiuqin and Luo, Xin and Xu, Zeshui},
	journal={IEEE Transactions on Intelligent Transportation Systems},
	year={2024},
	publisher={IEEE}
}

@article{yang2024latent,
	title={Latent factor analysis model with temporal regularized constraint for road traffic data imputation},
	author={Yang, Hengshuo and Lin, Mingwei and Chen, Hong and Luo, Xin and Xu, Zeshui},
	journal={IEEE Transactions on Intelligent Transportation Systems},
	year={2024},
	publisher={IEEE}
}

@article{luo2021alternating,
	title={An alternating-direction-method of multipliers-incorporated approach to symmetric non-negative latent factor analysis},
	author={Luo, Xin and Zhong, Yurong and Wang, Zidong and Li, Maozhen},
	journal={IEEE Transactions on Neural Networks and Learning Systems},
	volume={34},
	number={8},
	pages={4826--4840},
	year={2021},
	publisher={IEEE}
}

@article{luo2022neulft,
	title={NeuLFT: A novel approach to nonlinear canonical polyadic decomposition on high-dimensional incomplete tensors},
	author={Luo, Xin and Wu, Hao and Li, Zechao},
	journal={IEEE Transactions on Knowledge and Data Engineering},
	volume={35},
	number={6},
	pages={6148--6166},
	year={2022},
	publisher={IEEE}
}

@article{bi2023proximal,
	title={Proximal alternating-direction-method-of-multipliers-incorporated nonnegative latent factor analysis},
	author={Bi, Fanghui and Luo, Xin and Shen, Bo and Dong, Hongli and Wang, Zidong},
	journal={IEEE/CAA Journal of Automatica Sinica},
	volume={10},
	number={6},
	pages={1388--1406},
	year={2023},
	publisher={IEEE}
}

@article{yuan2025proportional,
	title={A proportional integral controller-enhanced non-negative latent factor analysis model},
	author={Yuan, Ye and Lu, Siyang and Luo, Xin},
	journal={IEEE/CAA Journal of Automatica Sinica},
	volume={12},
	number={6},
	pages={1246--1259},
	year={2025},
	publisher={IEEE}
}

@article{liao2025local,
	title={Local search-based anytime algorithms for continuous distributed constraint optimization problems},
	author={Liao, Xin and Hoang, Khoi and Luo, Xin},
	journal={IEEE/CAA Journal of Automatica Sinica},
	volume={12},
	number={1},
	pages={288--290},
	year={2025},
	publisher={IEEE}
}

@article{wu2024fine,
	title={A fine-grained regularization scheme for nonnegative latent factorization of high-dimensional and incomplete tensors},
	author={Wu, Hao and Qiao, Yan and Luo, Xin},
	journal={IEEE Transactions on Services Computing},
	year={2024},
	publisher={IEEE}
}

@article{chen2024efficient,
	title={An efficient industrial robot calibrator with multiplaner constraints},
	author={Chen, Tinghui and Yang, Weiyi and Zhang, Zhetao and Luo, Xin},
	journal={IEEE Transactions on Industrial Informatics},
	year={2024},
	publisher={IEEE}
}

@article{yang2024data,
	title={Data driven vibration control: A review},
	author={Yang, Weiyi and Li, Shuai and Luo, Xin},
	journal={IEEE/CAA Journal of Automatica Sinica},
	volume={11},
	number={9},
	pages={1898--1917},
	year={2024},
	publisher={IEEE}
}

@article{yang2021machine,
	title={A machine-learning prediction method of lithium-ion battery life based on charge process for different applications},
	author={Yang, Yixin},
	journal={Applied Energy},
	volume={292},
	pages={116897},
	year={2021},
	publisher={Elsevier}
}

@article{yuan2024fuzzy,
	title={A fuzzy PID-incorporated stochastic gradient descent algorithm for fast and accurate latent factor analysis},
	author={Yuan, Ye and Li, Jinli and Luo, Xin},
	journal={IEEE Transactions on Fuzzy Systems},
	volume={32},
	number={7},
	pages={4049--4061},
	year={2024},
	publisher={IEEE}
}

@article{yuan2024adaptive,
	title={Adaptive divergence-based non-negative latent factor analysis of high-dimensional and incomplete matrices from industrial applications},
	author={Yuan, Ye and Luo, Xin and Zhou, MengChu},
	journal={IEEE Transactions on Emerging Topics in Computational Intelligence},
	volume={8},
	number={2},
	pages={1209--1222},
	year={2024},
	publisher={IEEE}
}

@article{wang2021adaptive,
	title={Adaptive sliding window LSTM NN based RUL prediction for lithium-ion batteries integrating LTSA feature reconstruction},
	author={Wang, Zhuqing and Liu, Ning and Guo, Yangming},
	journal={Neurocomputing},
	volume={466},
	pages={178--189},
	year={2021},
	publisher={Elsevier}
}

@article{zhong2024alternating,
	title={Alternating-direction-method of multipliers-based adaptive nonnegative latent factor analysis},
	author={Zhong, Yurong and Liu, Kechen and Gao, Shangce and Luo, Xin},
	journal={IEEE Transactions on Emerging Topics in Computational Intelligence},
	year={2024},
	publisher={IEEE}
}

@article{zhang2020deep,
	title={Deep learning-based prognostic approach for lithium-ion batteries with adaptive time-series prediction and on-line validation},
	author={Zhang, Wei and Li, Xiang and Li, Xu},
	journal={Measurement},
	volume={164},
	pages={108052},
	year={2020},
	publisher={Elsevier}
}

@article{catelani2021remaining,
	title={Remaining useful life estimation for prognostics of lithium-ion batteries based on recurrent neural network},
	author={Catelani, Marcantonio and Ciani, Lorenzo and Fantacci, Romano and Patrizi, Gabriele and Picano, Benedetta},
	journal={IEEE Transactions on Instrumentation and Measurement},
	volume={70},
	pages={1--11},
	year={2021},
	publisher={IEEE}
}

@article{li2020state,
	title={State-of-health estimation and remaining useful life prediction for the lithium-ion battery based on a variant long short term memory neural network},
	author={Li, Penghua and Zhang, Zijian and Xiong, Qingyu and Ding, Baocang and Hou, Jie and Luo, Dechao and Rong, Yujun and Li, Shuaiyong},
	journal={Journal of power sources},
	volume={459},
	pages={228069},
	year={2020},
	publisher={Elsevier}
}

@article{wang2024distributed,
	title={A distributed adaptive second-order latent factor analysis model},
	author={Wang, Jialiang and Li, Weiling and Luo, Xin},
	journal={IEEE/CAA Journal of Automatica Sinica},
	volume={11},
	number={11},
	pages={2343--2345},
	year={2024},
	publisher={IEEE}
}

@article{ren2020data,
	title={A data-driven auto-CNN-LSTM prediction model for lithium-ion battery remaining useful life},
	author={Ren, Lei and Dong, Jiabao and Wang, Xiaokang and Meng, Zihao and Zhao, Li and Deen, M Jamal},
	journal={IEEE Transactions on Industrial Informatics},
	volume={17},
	number={5},
	pages={3478--3487},
	year={2020},
	publisher={IEEE}
}

@article{ma2023two,
	title={A two-stage integrated method for early prediction of remaining useful life of lithium-ion batteries},
	author={Ma, Guijun and Wang, Zidong and Liu, Weibo and Fang, Jingzhong and Zhang, Yong and Ding, Han and Yuan, Ye},
	journal={Knowledge-Based Systems},
	volume={259},
	pages={110012},
	year={2023},
	publisher={Elsevier}
}

@article{li2023state,
	title={State-of-health and remaining-useful-life estimations of lithium-ion battery based on temporal convolutional network-long short-term memory},
	author={Li, Chaoran and Han, Xianjie and Zhang, Qiang and Li, Menghan and Rao, Zhonghao and Liao, Wei and Liu, Xiaori and Liu, Xinjian and Li, Gang},
	journal={Journal of Energy Storage},
	volume={74},
	pages={109498},
	year={2023},
	publisher={Elsevier}
}

@article{li2023saliency,
	title={Saliency-aware dual embedded attention network for multivariate time-series forecasting in information technology operations},
	author={Li, Jiajia and Tan, Feng and He, Cheng and Wang, Zikai and Song, Haitao and Hu, Pengwei and Luo, Xin},
	journal={IEEE Transactions on Industrial Informatics},
	volume={20},
	number={3},
	pages={4206--4217},
	year={2023},
	publisher={IEEE}
}
\end{document}